# From Human-Computer Interaction to Human-Robot Social Interaction

Tarek Toumi and Abdelmadjid Zidani

LaSTIC Laboratory, Computer Science Department
University of Batna, 05000 Algeria

**Abstract**
Human-Robot Social Interaction became one of active research fields in which researchers from different areas propose solutions and directives leading robots to improve their interactions with humans. In this paper we propose to introduce works in both human robot interaction and human computer interaction and to make a bridge between them, i.e. to integrate emotions and capabilities concepts of the robot in human computer model to become adequate for human robot interaction and discuss challenges related to the proposed model. Finally an illustration through real case of this model will be presented.

*Keywords*: *Human-Computer Interaction, Human-robot social interaction, social robot.*

## 1. Introduction

Building robots that can interact socially and in a robust way with humans is the main goal of Human-robot social interaction researchers, a survey of previous works could be found in [1], [2], [3]. However until today there is no robot that can interact efficiently with humans, this because of the enormous capacity of human to interact socially and also the difficulties to understand how human do it. This leads to the need of cooperation of researchers with different backgrounds: in psychology, in human computer interaction, in robotics, etc.

Works in physiology of emotion provides indirect indices on the emotional state of human, however it is difficult to use these indices. Using an analytical model to represent emotional state of human has the disadvantages of not being able to model true felt emotions. Indeed, this solution has limitations, because our poor knowledge of conditions and emergence processes of emotions, but also the degree of reduction that we will have to do to simulate emotions, however a believe that despite the gap between likely simulated emotions and the emotions actually felt by human, an analytical model can emotionally engage the user in his interactions and produce a feeling of presence in the device.

Emotional models among the most commonly cited, we find that of Trappl et al [4], and that of Ortony et al [5] also called OCC model. Those models are based on theories of evaluations that allow specifying properties of critical events that can cause particular emotions [6], [7], [8].

Concerning Works in robotics Bartneck [9] proposed a definition and design guidelines of social robots. Several architectures and theories have been proposed [5], [4], and some of them have been used in the design step. Many others works could be find in [1], [2], [3].

In which concern the works in human computer interaction, Norman [10] by his action theory, describes how humans doing their actions by modeling cognitive steps used in accomplishment a task. Jean Scholtz [11] used Norman's action theory and described an evaluation methodology based on situational awareness.

From typical models used in human computer interaction, this paper proposes some modifications to be used in social interaction between robot and human, i.e. to integrate emotions and capabilities concepts of the robot in human computer interaction model to become adequate for human robot interaction. This model is based on a modification of action theory proposed by Norman [10]. Finally an illustration through real case of this model will be presented.

## 2. Human Computer Interaction

Knowing that the first International Conference on Human–Robot Interaction was sponsored by ACM's Computer–Human Interaction Special Interest Group [2] this shows how many contributions from HCI organizations to develop human robot interaction. HRI benefits in methodologies, design principles, and computing metaphors from HCI contributions. Due to the vast work in Human Computer interaction, we present only a brief survey of most used models. More reviews on human computer interaction could be found in [12].

A model is a simplification of reality. HCI models provide simplification of the complex interaction between humans and computers [13]. Different models of Human-Computer Interaction (HCI) were proposed in the literature : the model most commonly used , named interaction model of Norman or action theory, proposed by D. Norman [10], it is used to model different cognitive steps of task performing, we will describe this model in detail later.
We also provide a small description of different models used to model HCI:

- Instrumental interaction model [14], is based on the notion of interaction instrument, acting intermediary between the user and the manipulated objects. In this model, manipulated objects and then the instruments are defined, and finally the principle of reification to create new objects and new instruments.
- Physical interaction model of SATO [15]: To take into account the real objects, physical interaction model of SATO was proposed. The purpose of SATO model is to model interactive systems and allows to mix real world elements or "physical space," and elements of the digital world called "media space" or "virtual world" .
- Fitts' model [13]: is a predictive model, it models the information-processing capability of the human motor system.
- Guiards' model [13]: is a descriptive model, it models bimanual control, i.e. How humans use their hands in everyday tasks, such as writing, manipulating objects, etc.
- Model of Seeheim: The architecture of Seeheim model was the first to establish a clear separation between interface and the functional part. This interface consists of three functional units: The Functional Core: which includes the concept and functions of the area. The Presentation has main role to make perceptible the status of domain concepts and allow manipulation by the user. Controller Dialogue which serves as a bridge between Functional Core and Presentation.
- Arch reference model: or revised Seeheim model is a refinement of the Seeheim model mentioned before. There are five components organized in the form of an arch: Functional Core, Adapter Domain, Dialogue Controller, Component Presentation and finally, interaction Component.
- Multi-agents models: many models within multi-agent approach has been proposed: MVC (Model-View-Controller) model, PAC (Presentation-Abstraction-Control) and AMF model (Agents multi-faceted). Note that all these models are based on the concept of software agents that communicate with each other.

## 3. Norman's Model

The action theory proposed by Norman [10] allows modeling the various cognitive steps used in performing a task. It models the process of execution and evaluation into seven stages as following (Fig.1):

- Formulation of the goal: thinking in high level terms of what do we want to accomplish.
- Formulation of the intention: thinking more specifically about what will satisfy this goal.
- Specification of the action: determining what actions are necessaries to carry out the intention. These actions will then be carried out one at a time.
- Execution of the action: physically doing the action. In computer terms this would be selecting the commands needed to carryout a specific action.
- Perception of the system state: the user should then assess what has occurred based on the action specified and execution. In the perception part the user must notice what happened.
- Interpretation of the system state: having perceived the system state, the user must now use his knowledge of the system to interpret what happened.
- Evaluation of the outcome: the user now compares the system state (as perceived and interpreted by himself) to the intention and to decide if progress is being made and what action will be needed next.

These seven stages are iterated until the intention and goal are achieved, or the user decides that the intention or goal has to be modified.

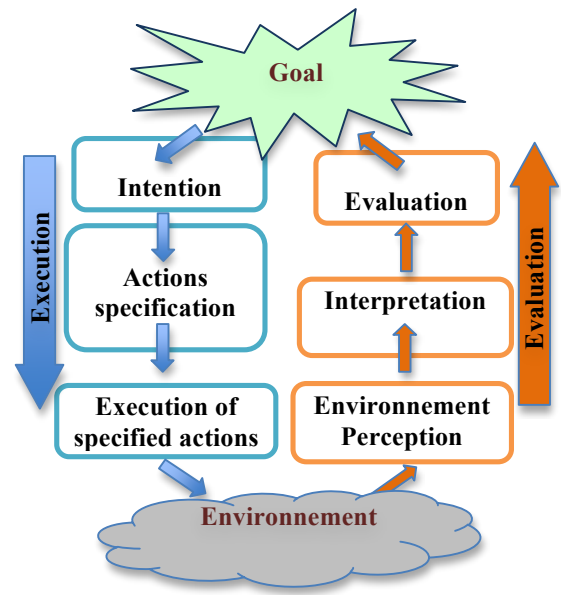

Fig.1: Seven stages of the Norman model

## 4. Human-Robot Interaction

Study of interactions between humans and robots, is the main concern of HRI. Different modalities of HRI could be found in the reality: Human-robot Visual interaction, Human-robot vocal Interaction and Human robot Social interaction, etc.

### 4.1 Human-robot Visual interaction

One of the issues in this context is to develop robots capable to visually interact with humans. Vision system is an important part in human robot visual interaction, where robot can identify geometric shapes and different colors [16], or classifying images [17], or face detection and trucking human in natural cluttered environment [18], or recognizing basic humans gestures. Many works [19], [20], [21], [22] have been focused on this topic.

### 4.2 Human-robot vocal Interaction

Speech conversation is the natural social interaction between humans [23]. Research in this area addresses the integration of speech recognition in the vocal interactive interface of service robots, e.g. it is important that a guide robot initiates a dialogue with visitors in museum, and provides the appropriate services, like showing parts of the museum, Etc.

### 4.3 Human robot Social interaction

To be accepted in human environment, it is important that robots adopt social behaviors through emotional facial expressions, and respect social conventions [1], [24], [25].
Many studies have been made involving such emotional facial expressions like Avatars in the virtual anthropomorphic appearance, often implemented by Conversational Animated Agents and interact with humans or other agents through a multimodal communicative behavior [26], for example Greta [26], [27], REA [28], Grace or Cherry [29]. Cherry is mounted on a robot and orients people in University of Central Florida.
Regarding physical robots, two kinds have been designed for social interaction: zoomorphic robots and anthropomorphic robots [1]. Zoomorphic robots are mostly toys with the appearance of cats or dogs, for example Aibo (Sony) or iCat (Philips).
Anthropomorphic robots are more dedicated to social interaction and need to express their internal emotional states, goals and their desires.
Many works have been done to develop such robots, e.g., iCat or Leonardo [30].

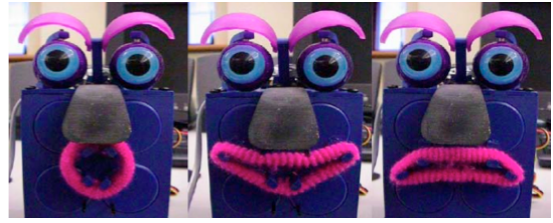

Fig. 2: Facial expressions of ESRA from Robodyssey Systems LLC.

## 5. From HCI to Human Robot Social Interaction

We consider that the human robot interaction and human computer interaction can benefit from each other progress, even robots in human robot interaction are often designed to be mobile and autonomous which differ from computers.

In the following parts we give an example of the use of HCI model to model HRI by adapting the action theory described before, with an integration of emotions and robots capabilities in the process. The modified model (Fig. 3) become:

- Formulation of the goal.
- Formulation of the intention.
- Verification of the robot capability according to the intentions.
- Specification of actions sequence, ensuring the goal establishment.
- Simulation of robot emotions emerges from a relationship with objects. For example; the robot is happy about something, angry against something, satisfied about something, etc.
- Execution of the action specified by actions sequence and emotions simulated.
- Perception of the environments and emotions of human after the completion of the task by the robot.
- Interpretation of the perceived status of the system according to goal.
- Evaluation of the system state in terms of goals and intentions. This often leads to a new set of goals and intentions.

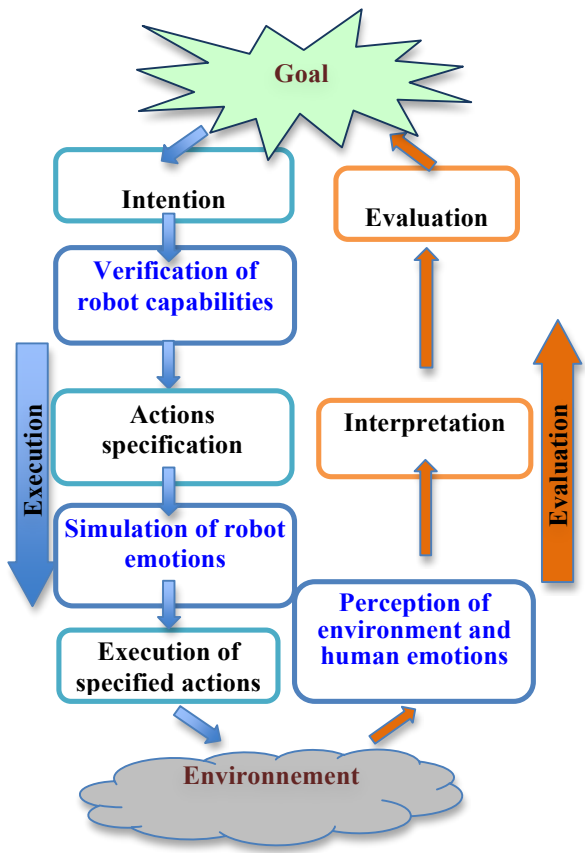

Fig.3: Norman Model adapted

- Simulation of robot emotions: In this example it is assumed that robot is in good humor and doing its normal task.
- Execution: the robot executes the prior actions plan to change the system state.
- Perception of the environment and the human emotions: the robot perceives that the human shakes hands with his friend.
- Interpretation: it interprets that the human is perturbed.
- Evaluation: the robot values that the task is temporarily suspended and that it must drop the bottle. This creates a new set of goals and intentions represented by the repetition of the task as soon as human will be free.

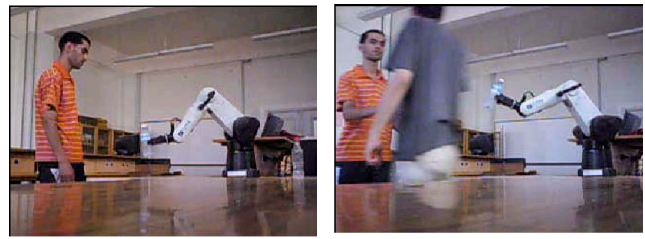

(a)  (b)

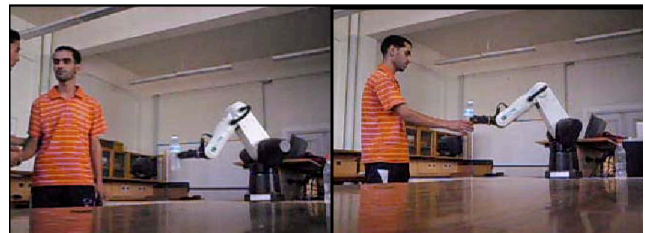

(c)  (d)

Fig.4: Social behavior of the robot.

## 6. Practical case

The following section describes an illustration of the use of adapted action theory in practical case.

This example introduces social behavior of the robot: the robot plans to give a bottle to the human (Fig.4.a) however the human is disturbed by another human (Fig.4.b). The robot temporarily suspends the task and drops the bottle (Fig.4.c). When the human become free the robot automatically repeats the task (Fig.4.d).

The adapted model applied for this case is:

- Formulation of the goal: To give a bottle to human.
- Formation of the intention: Intention to assist human.
- Verification of the robot capability: the robot has all necessaries capabilities for the effective realization of the task
- Specification of an action plan: the action plan is the following: detection of human position, trajectory planning and movement generation.

## 7. Practical Challenges

To build robots that can interact socially with humans we have first to understand how humans interact between each other, even human has the capacity to interact with his environment very easily, we do not know until now how to emulate his behavior. The following parts describe the main three parts of our previous model: Verification of the robot capability, Simulation of robot emotions and perception of human emotions and trying to make a bridge between how human do it and how robot should emulate it.

### 7.1. Verification of the robot capability

Human can predict his capabilities in most of situations. For example if someone is asked to drive a car, he can say that he can not do it, even before to try to drive it because he knows his capacities. According to the experiments of Dr. Daryl J. Bem, we have the ability to predict future events [31].

This characteristic that we call here Knowledge of personal abilities (Possibility of achieving the goal) could be fellows two ways: From previous experiences and from availability of the elements that lead to achieve the goal. Concerning previous experiences: is that the future is determined by the past. If we want to predict the future, we should know the laws that determine the future based on its past. we can predict the possibility of achieving a goal from our previous experiences, we may know for example that we can change the brakes of our car if we have previous successful experiences of changing brakes of the same car , and vice versa we known that we are not able to do it if we failed in changing the brakes before. Concerning the case of robot we can make a data base in which robot save all its previous goals and results (successful or no) to achieve them. Also we can predict the possibility of achieving the goal if we have all necessaries elements to do it, for example, a person can predict that he can write a latter if he has all elements like: paper, pen, knowledge of writing, etc., and vice versa he knows that he is not able to write a latter because of the absence of some or all the above elements. This could be memorized in the data base of robot in which all necessaries elements of achievement of previous goal should be saved.

7.2. Simulation of robot emotions: Execution of expressions and feelings:

Social aspect is the most important characteristic of human. We have an enormous capacity to express our feelings through facial and physical expressions, as well as our abilities to understand these expressions.

Facial expressions are considered as key means to express our feelings with a complex facial muscles movements, resulting in a whole emotional expression like happiness, etc. Body also could be used in expression of human feelings, for example, when someone wins the race, he raise his hands reflecting his joy and pride, as well as body shivering indicates extreme cold or fear. Sometimes human hide his real emotions and does not shows the reality of his feeling, such as pretending to laugh, although he does not wish to do so. In the literature of robotics there are many robots that can generate emotional expressions.

7.3. Perception of others' emotional expressions:

Reading the facial or physical expressions of others has a close relationship with memory, Knowing a persons' personality helps to read his physical and facial expressions more accurate by linking them with previous expresions in the memory. Robot have to be sensitive to human emotions and can inderstand human social signals to be socially accepted by him. Also there are many works in the literature on the interpretation of emotional expressions of human. Chastagnol et al [32] focused on the steps necessery to build the emotion detection module of a spoken dialog system for a social robotic companion.

## 8. Conclusion and future works

We have shown how to use human computer interaction model in the case of human robot social interaction, this allows us to benefit from progress realized by the HCI community. This could be a starting point for many others adaptations of HCI models like model of SATO, instrumental model, multi agent models, etc.

**Acknowledgements**

Many thanks to A. Hafid and to Professors in the department of Industrial Engineering in the University of Batna for their experimental support.

## References
[1] T. Fong, I. Nourbakhsh, and K. Dautenhahn, "A survey of socially interactive robots," Robotics and autonomous systems, vol. 42, pp. 143-166, 2003.
[2] M. A. Goodrich and A. C. Schultz, "Human-robot interaction: a survey," Foundations and Trends in Human-Computer Interaction, vol. 1, pp. 203-275, 2007.
[3] J. Broekens, M. Heerink, and H. Rosendal, "Assistive social robots in elderly care: a review," Gerontechnology, vol. 8, pp. 94-103, 2009.
[4] R. Trappl, P. Petta, and S. Payr, Emotions in humans and artifacts: The MIT Press, 2002.
[5] A. Ortony, The cognitive structure of emotions: Cambridge university press, 1990.
[6] R. S. Lazarus, Emotion and adaptation: Oxford University Press New York, 1991.
[7] I. J. Roseman, "Appraisal determinants of emotions: Constructing a more accurate and comprehensive theory," Cognition & Emotion, vol. 10, pp. 241-278, 1996.
[8] K. R. Scherer, "Criteria for emotion-antecedent appraisal: A review," in Cognitive perspectives on emotion and motivation, ed: Springer, 1988, pp. 89-126.
[9] C. Bartneck and J. Forlizzi, "A design-centred framework for social human-robot interaction," in Robot and Human Interactive Communication, 2004. ROMAN 2004. 13th IEEE International Workshop on, 2004, pp. 591-594.
[10] D. A. Norman and S. W. Draper, User centered system design; new perspectives on human-computer interaction: L. Erlbaum Associates Inc., 1986.
[11] J. Scholtz, "Theory and evaluation of human robot interactions," in System Sciences, 2003. Proceedings of the

36th Annual Hawaii International Conference on, 2003, p. 10 pp.
[12] M. Turk, "Multimodal interaction: A review," Pattern Recognition Letters, vol. 36, pp. 189-195, 2014.
[13] J. M. Carroll, HCI models, theories, and frameworks: Toward a multidisciplinary science: Morgan Kaufmann, 2003.
[14] M. Beaudouin-Lafon, "Instrumental interaction: an interaction model for designing post-WIMP user interfaces," in Proceedings of the SIGCHI conference on Human factors in computing systems, 2000, pp. 446-453.
[15] K. Sato and Y.-k. Lim, "Physical interaction and multi-aspect representation for information intensive environments," in Robot and Human Interactive Communication, 2000. RO-MAN 2000. Proceedings. 9th IEEE International Workshop on, 2000, pp. 436-443.
[16] M. El Sayed, M. H. Awadalla, H. E. I. Ali, and R. F. Mostafa, "Interactive Learning for Humanoid Robot," IJCSI International Journal of Computer Science Issues, Vol. 9, Issue 4, No 1, 2012.
[17] M. S. Lotfabadi, "The Presentation of a New Method for Image Distinction with Robot by Using Rough Fuzzy Sets and Rough Fuzzy Neural Network Classifier," IJCSI International Journal of Computer Science Issues, Vol. 8, Issue 4, No 1, July 2011.
[18] M. Rahat, M. Nazari, A. Bafandehkar, S. S. Ghidary, I. Marand, and I. Dezful, "Improving 2D Boosted Classifiers Using Depth LDA Classifier for Robust Face Detection," IJCSI, International Journal of Computer Science Issues, Vol. 9, Issue 3, No 2, May 2012
[19] L.-P. Morency, C. M. Christoudias, and T. Darrell, "Recognizing gaze aversion gestures in embodied conversational discourse," in Proceedings of the 8th international conference on Multimodal interfaces, 2006, pp. 287-294.
[20] D. Demirdjian, "Combining geometric-and view-based approaches for articulated pose estimation," in Computer Vision-ECCV 2004, ed: Springer, 2004, pp. 183-194.
[21] K. Nickel and R. Stiefelhagen, "Visual recognition of pointing gestures for human–robot interaction," Image and Vision Computing, vol. 25, pp. 1875-1884, 2007.
[22] J. Triesch and C. Von Der Malsburg, "A gesture interface for human-robot-interaction," in Automatic Face and Gesture Recognition, 1998. Proceedings. Third IEEE International Conference on, 1998, pp. 546-551.
[23] D. Feil-Seifer and M. J. Mataric, "Defining socially assistive robotics," in Rehabilitation Robotics, 2005. ICORR 2005. 9th International Conference on, 2005, pp. 465-468.
[24] J. Gratch and S. Marsella, "Lessons from emotion psychology for the design of lifelike characters," Applied Artificial Intelligence, vol. 19, pp. 215-233, 2005.
[25] R. W. Picard, A. Computing, and M. Editura, "MIT Press," Cambridge, MA, 1997.
[26] I. Poggi, C. Pelachaud, F. de Rosis, V. Carofiglio, and B. De Carolis, "Greta. a believable embodied conversational agent," in Multimodal intelligent information presentation, ed: Springer, 2005, pp. 3-25.
[27] C. Pelachaud and M. Bilvi, "Computational model of believable conversational agents," in Communication in multiagent systems, ed: Springer, 2003, pp. 300-317.
[28] J. Cassell, T. Bickmore, M. Billinghurst, L. Campbell, K. Chang, H. Vilhjálmsson, and H. Yan, "Embodiment in conversational interfaces: Rea," in Proceedings of the SIGCHI conference on Human factors in computing systems, 1999, pp. 520-527.
[29] C. L. Lisetti, S. M. Brown, K. Alvarez, and A. H. Marpaung, "A social informatics approach to human-robot interaction with a service social robot," Systems, Man, and Cybernetics, Part C: Applications and Reviews, IEEE Transactions on, vol. 34, pp. 195-209, 2004.
[30] A. L. Thomaz and C. Breazeal, "Robot learning via socially guided exploration," in Proceedings of the 6th Intl. Conf. on Development and Learning, London, 2007.
[31] D. J. Bem, "Feeling the future: experimental evidence for anomalous retroactive influences on cognition and affect," Journal of personality and social psychology, vol. 100, p. 407, 2011.
[32] C. Chastagnol, C. Clavel, M. Courgeon, and L. Devillers, "Designing an emotion detection system for a socially intelligent human-robot interaction," in Natural Interaction with Robots, Knowbots and Smartphones, ed: Springer, 2014, pp. 199-211.

**Tarek TOUMI** received his master degree in robotics in 2004, he is an assistant Professor in the university of Batna since 2004. He is actually doing his PhD in computer science, His current research interests focus on human robot interaction(HRI), Social Robots, etc.

**Abdelmadjid Zidani** has a professor degree in computer science. He received his Ph.D degree in computer science from the University of Batna since 2002. He is a professor in computer science since 2011. His current research interests focus on human computer interaction (HCI), computer supported collaborative work (CSCW), etc